\documentclass[runningheads]{llncs}
\usepackage{times}
\usepackage{epsfig}
\usepackage{subfig}
\usepackage{graphicx}
\usepackage{amsmath}
\usepackage{amssymb}

\usepackage{booktabs}
\usepackage{colortbl}
\usepackage[inline]{enumitem}
\usepackage{algorithm}
\usepackage[noend]{algpseudocode}
\usepackage{mathtools}
\usepackage{tikz}
\usetikzlibrary{matrix}
\usetikzlibrary{positioning}
\usetikzlibrary{calc}
\usetikzlibrary{fit}
\usetikzlibrary{shapes}

\usepackage{color}
\usepackage[width=122mm,left=12mm,paperwidth=146mm,height=193mm,top=12mm,paperheight=217mm]{geometry}

\usepackage[pagebackref=true,breaklinks=true,letterpaper=true,colorlinks,bookmarks=false]{hyperref}

\begin{document}

\pagestyle{headings}
\mainmatter
\def\ECCV16SubNumber{818}  

\title{Radiometric Scene Decomposition: \\Scene Reflectance, Illumination, and Geometry from RGB-D Images}

\titlerunning{Radiometric Scene Decomposition}

\authorrunning{Stephen Lombardi and Ko Nishino}
\author{Stephen Lombardi \and
Ko Nishino
}
\institute{Department of Computer Science\\
Drexel University, Philadelphia, PA\\
{\tt\small \{sal64, kon\}@drexel.edu}}

\maketitle

\begin{abstract}

Recovering the radiometric properties of a scene (i.e., the reflectance, illumination, and geometry) is a long-sought ability of computer vision that can provide invaluable information for a wide range of applications. Deciphering the radiometric ingredients from the appearance of a real-world scene, as opposed to a single isolated object, is particularly challenging as it generally consists of various objects with different material compositions exhibiting complex reflectance and light interactions that are also part of the illumination. We introduce the first method for radiometric scene decomposition that handles those intricacies. We use RGB-D images to bootstrap geometry recovery and simultaneously recover the complex reflectance and natural illumination while refining the noisy initial geometry and segmenting the scene into different material regions. Most important, we handle real-world scenes consisting of multiple objects of unknown materials, which necessitates the modeling of spatially-varying complex reflectance, natural illumination, texture, interreflection and shadows. We systematically evaluate the effectiveness of our method on synthetic scenes and demonstrate its application to real-world scenes. The results show that rich radiometric information can be recovered from RGB-D images and demonstrate a new role RGB-D sensors can play for general scene understanding tasks.

\end{abstract}

\section{Introduction}

Radiometric decomposition, the problem of recovering radiometric properties, namely the reflectance, illumination, and geometry of objects and scenes from images, is a challenging but important task in computer vision, which finds applications in a wide range of areas. The recovered radiometric properties can provide invaluable clues about the scene that are otherwise inaccessible. For instance, the reflectance of an object can tell what material it is made of \cite{Wang_CVPR09,Gu_CVPR12} and the recovered illumination can provide a snapshot of the surrounding environment which can tell us the location. The estimated radiometric properties can benefit autonomous navigation and grasp planning for robotics, appearance prediction for tracking and motion estimation in computer vision, and realistic image synthesis in graphics.

Past works on radiometric decomposition have focused on isolated single objects \cite{Lombardi_PAMI15} or have heavily relied on simplistic assumptions about the reflectance and illumination, such as Lambertian reflectance and point light sources \cite{Barron_CVPR13,Yu_CVPR13,Or-el_CVPR15}. In contrast, radiometric decomposition of real-world scenes poses unique challenges stemming from the various factors that contribute to the complex scene appearance. These include the complex reflectance of real-world surfaces which can have textures as well, the inter-object light transport including interreflection and shadows, and the complex material composition of the scene and individual objects.

In this paper, we introduce the first method for radiometric scene decomposition that handles the intricate factors underlying real-world scene appearance. For this, we leverage RGB-D images to bootstrap geometry recovery. The goal is to simultaneously recover the complex reflectance and natural illumination while refining the noisy initial geometry and segmenting the scene into different material regions. Most important, we want to handle real-world scenes consisting of multiple objects of unknown materials (i.e., not just single objects and Lambertian reflectance), which necessitates the modeling of complex reflectance and illumination as well as non-local light interaction. For this, we derive a Bayesian formulation and introduce scene representations and prior distributions that resolve the ambiguities among the radiometric scene properties and exploit their latent regularities. The paper makes four key contributions based on this Bayesian formulation:

\begin{itemize}
        \item \textbf{Spatially-varying reflectance} is modeled by decoupling it into distinct isotropic BRDFs and diffuse texture. We derive a spatial segmentation model that dissects the scene into regions of complex but distinct reflectance while simultaneously allowing diffuse texture to vary within the regions. This leads to a compact representation of complex real-world scene reflectance.

\item \textbf{Interreflection and shadows} are modeled in the image formation likelihood by fully simulating the indirect illumination with path tracing rendering \cite{Kajiya_SG86}. We also derive an algorithm for computing the gradients of the log posterior with respect to reflectance and illumination to enable tractable MAP estimation.

    \item \textbf{Geometry refinement and scene segmentation} are achieved by leveraging a set of smoothly varying geometric bases. These geometric bases are defined on the scene surface and enable refinement of low-frequency errors characteristic of RGB-D sensors due to interreflection and oversmoothing with a small set of parameters.

    \item \textbf{Full radiometric scene decomposition} is achieved through maximum a posterior estimation with careful variable initialization and ordered optimization. We begin by initializing the segmentation of the scene by decomposing the geometry into contiguous clusters with similar curvature and color and incorporate each variable into the optimization one at a time.
\end{itemize}

We evaluate the effectiveness of our framework on a novel set of synthetic and real scenes. We conduct an extensive quantitative accuracy evaluation on a synthetic image set using shapes with artificially perturbed geometry and rendered with measured reflectance under natural illumination environments. We also qualitatively evaluate our method on a set of synthetic and real-world scenes. The results show that radiometric scene decomposition can be reliably computed from RGB-D images, which gives completely new roles for RGB-D data to play in computer vision and provides novel tools for tackling challenging scene understanding tasks.

\section{Related Work}

A few methods have been proposed in the past that concern the recovery of radiometric information from RGB-D images. Or-El et al. fuse depth and color information from multiple RGB-D images to estimate high resolution geometry \cite{Or-el_CVPR15}. This is achieved by refining the geometry captured in the RGB-D image with shape-from-shading. Barron and Malik recover Lambertian reflectance, spatially-varying illumination, while also refining geometry from a single RGB-D image \cite{Barron_CVPR13} with a large-scale energy minimization using several unique priors. Yu et al. simultaneously estimate albedo and illumination from RGB-D by explicitly refining surface normals \cite{Yu_CVPR13}. These methods all heavily rely on the simplistic assumption of Lambertian reflectance and do not model the light interaction among scene elements.

Other methods attempt to estimate radiometric object (not scene) properties given very accurate geometry. Goldman et al. estimate geometry and spatially-varying BRDFs from a set of images under known point-light illumination \cite{Goldman_IJCV10}. Similar to our approach, the authors note that objects are typically composed of a small number of materials. They estimate each of these ``fundamental'' materials and a material weight map that controls the mixture of them at each pixel. Our approach to modeling scene reflectance is similar but has several critical distinctions. First, the segmentation of our scene is imposed over the 3D geometry rather than in the image space to model multiple views of a scene. We also leverage a set of geometric bases to model the spatial segmentation that encourages contiguous material regions. Most important, we do not allow a mixture of fundamental materials (i.e., a surface point only exhibits one of $K$ base materials) and decouple the texture from reflectance, which we believe is a more faithful model of complex appearance of real-world scenes.

Lombardi and Nishino recover reflectance and natural illumination from a single image given known geometry of a uniform object in the real world \cite{Lombardi_ECCV12}. They solve this problem in a Bayesian formulation that leverages strong priors on real-world reflectance functions and statistical as well as information-theoretic priors on natural illumination. They demonstrate joint estimation of reflectance and illumination from isolated objects captured in the wild \cite{Lombardi_PAMI15}. Our work fully extends this approach to handle real-world scenes, which is nontrivial due to the complex factors underlying real-world scene appearance.

In graphics, there has been a body of work on data-driven modeling of object appearance. Recently, Dong et al. estimated spatially-varying reflectance and natural illumination from a video of a single object with known accurate geometry \cite{Dong_SIGGRAPH14}. Their method works by examining the change in appearance of an object as it rotates through space. The method requires many input images and is not robust to noisy geometry from RGB-D sensors. Most important, the method only considers single objects and is not applicable to real-world scenes that consist of multiple objects whose reflectance, geometry, and illumination conspire to give rise to the complex scene appearance. Karsch et al. estimate reflectance, illumination, and geometry from a single image \cite{Karsch_TOG14} by sequentially estimating each variable one at a time: first geometry, then reflectance, then illumination. In contrast, our method models the crosstalk between each of these variables which is critical to decompose the scene appearance resulting from the interdependence between each of these variable.

\section{Bayesian Radiometric Scene Decomposition}

We capture multiple images of a scene with an RGB-D sensor. In practice, as explained in Section \ref{sec:exp}, we use a pre-calibrated DLSR-RGB-D setup to capture high-dynamic range image RGB-D images. We formulate radiometric scene decomposition in a Bayesian framework where we model the complex spatially-varying reflectance ($\mathbf{R}$), natural illumination ($\mathbf{L}$), and geometry ($\mathbf{G}$) as random variables. Although we use RGB-D images that capture scene geometry, the geometry still needs to be estimated as it is noisy. In other words, the geometry extracted from the RGB-D images serve as an initial estimate of the scene geometry. We write the observations (i.e., the HDR pixel values) as $\mathbf{I}$ and use Bayes' rule (with the assumption that reflectance, illumination, and geometry are statistically independent) to write the posterior probability,
\begin{equation}
p\left(\mathbf{R}, \mathbf{L}, \mathbf{G} | \mathbf{I}\right) \propto
    p\left(\mathbf{I} | \mathbf{R}, \mathbf{L}, \mathbf{G}\right)
    p\left(\mathbf{R}\right)
    p\left(\mathbf{L}\right)
    p\left(\mathbf{G}\right).
\end{equation}
This allows us to specify prior distributions on the reflectance, illumination, and geometry that capture the real-world variation of the variables.

We write the likelihood function using a log-Laplace distribution centered on the predicted irradiance,
\begin{equation}
p(\mathbf{I} | \mathbf{R}, \mathbf{L}, \mathbf{G}) 
    \propto \exp \left[ -| \log \mathbf{I}_{\mathbf{p}} - \log \hat{\mathbf{I}}(\mathbf{p}; \mathbf{R}, \mathbf{L}, \mathbf{G})|\sigma^{-2} \right],
\end{equation}
where $\hat{\mathbf{I}}$ is a function that computes the predicted irradiance based on the current estimates of the reflectance, illumination, and geometry. We use the logarithm of the predicted irradiance because it makes the likelihood function independent of the scale of the intensity values.

From the next section, we explain each of the variables and priors in detail.

\section{Representing Scene Reflectance}

Modeling real-world scenes is difficult because real-world surfaces can exhibit complex reflectance that also varies arbitrarily along the surface. To exploit regularities commonly found in real-world surfaces, we model the scene as a composition of a handful of distinct materials each represented with a compact isotropic reflectance model. We develop a novel spatial segmentation model to estimate the spatial extent of each material region while also allowing diffuse texture to vary within material regions. For completeness, we first review the modeling of individual material reflectance.

\subsection{Representing Single Material Reflectance}

Faithfully modeling the angular behavior of complex, real-world reflectance is crucial to solving radiometric decomposition. For this, we turn to the Directional Statistics BRDF model \cite{Nishino_ICCV09,Nishino_JOSA11} that has been used in the past for recovering complex reflectance, natural illumination, and geometry \cite{Lombardi_CVPR12,Lombardi_ECCV12,Lombardi_PAMI15}. Past work has shown it is capable of accurately modeling real-world reflectance while also providing a compact parameterization amenable to probabilistic inference.

The DSBRDF model is written as a sum of reflectance lobes,
\small
\begin{equation}
f_{\lambda}(\theta_d, \theta_h) =
    \sum_r \left(\exp\left\{ \kappa_r(\theta_d) \cos^{\gamma_r(\theta_d)}(\theta_h) \right\} - 1\right) c_{r,\lambda},
\end{equation}
\normalsize
where $\kappa$ controls the intensity of the lobe as $\theta_d$ varies, $\gamma$ controls the acuteness of the reflectance lobe in the angular space as $\theta_d$ varies, and $c_{j,\lambda}$ controls the color of the lobe. Modeling the intensity and shape of each reflectance lobe as $\theta_d$ varies allows the DSBRDF to model many real-world materials.

As in previous work, we use the DSBRDF reflectance bases to model the functions
$\kappa(\theta_d)$ and $\gamma(\theta_d)$.
%
%
We can then compactly model real-world BRDFs with the coefficient vector $\mathbf{\Psi}$, which takes a linear combination of the reflectance bases (in practice we find 7 bases are sufficient). We utilize the DSBRDF prior distributions that carefully constrain the variability of real-world reflectance. In particular, we adopt the DSBRDF coefficient prior learned from the MERL database of BRDFs \cite{Lombardi_PAMI15}. We write the prior using a mixture of Gaussians,
\begin{equation}
p(\mathbf{\Psi}) = \sum_i \pi_i \mathcal{N}\left(\mathbf{\Psi} | \mu_i, \Sigma_i\right),
\end{equation}
where $\pi_i$, $\mu_i$, and $\Sigma_i$ are the parameters of the distribution after fitting to the MERL BRDF database. We also adopt the DSBRDF chromaticity prior that constrains the DSBRDF lobes to have similar hues \cite{Lombardi_PAMI15}. This is written using a von-Mises Fisher distribution on each pair of lobe chromaticities,
\begin{equation}
p(\mathbf{c}_{r}|\mathbf{c}_{r'}) \propto \exp\left[ \kappa_h \mathbf{h}^{\textsf{T}}(\mathbf{c}_{r}) \mathbf{h}(\mathbf{c}_{r'}) \right],
\end{equation}
where $\mathbf{h}$ computes the ``hue vector'' of the chromaticity \cite{Lombardi_PAMI15}.

\subsection{Segmenting the Scene by Material}

\begin{figure}[t]
\centering
    \begin{tikzpicture}[inner sep=0mm,scale=0.8,every node/.style={scale=0.8}]
        \matrix (table) [row sep=1mm, column sep=1mm, ampersand replacement=\&] {
            \node[rotate=90, align=center,inner sep=0mm] {\sf\scriptsize Geometric\\[-1mm]\sf\scriptsize Bases}; \&
            \node[rotate=0, align=center,inner sep=0mm] (A1) {\includegraphics[angle=90,width=15mm,trim=270 40 260 60,clip=true]{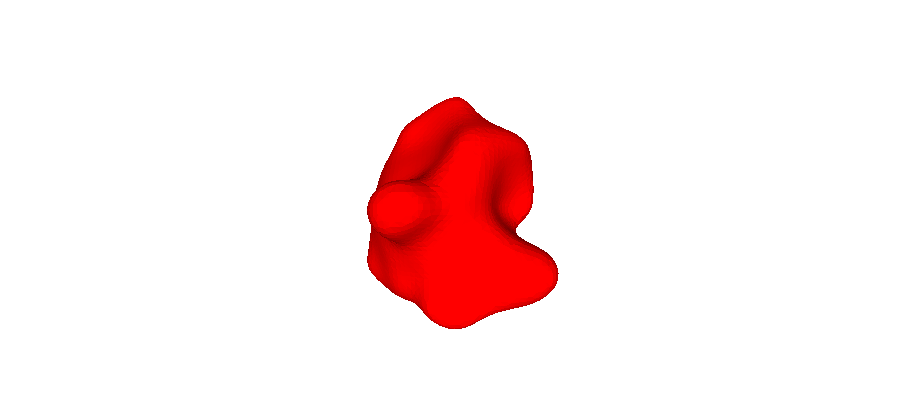}\\[-1mm]$\mathbf{g}^{(1)}$}; \&
            \node[rotate=0, align=center,inner sep=0mm] (A2) {\includegraphics[angle=90,width=15mm,trim=270 40 260 60,clip=true]{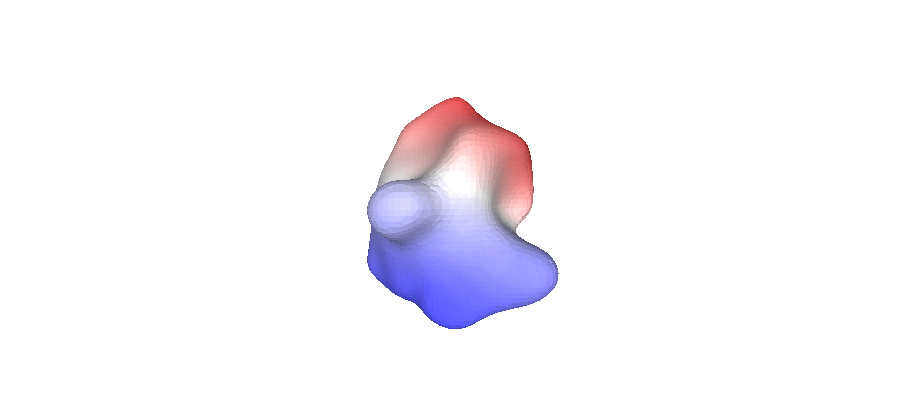}\\[-1mm]$\mathbf{g}^{(2)}$}; \&
            \node[rotate=0, align=center,inner sep=0mm] {\ldots}; \&
            \node[rotate=0, align=center,inner sep=0mm] (A3) {\includegraphics[angle=90,width=15mm,trim=270 40 260 60,clip=true]{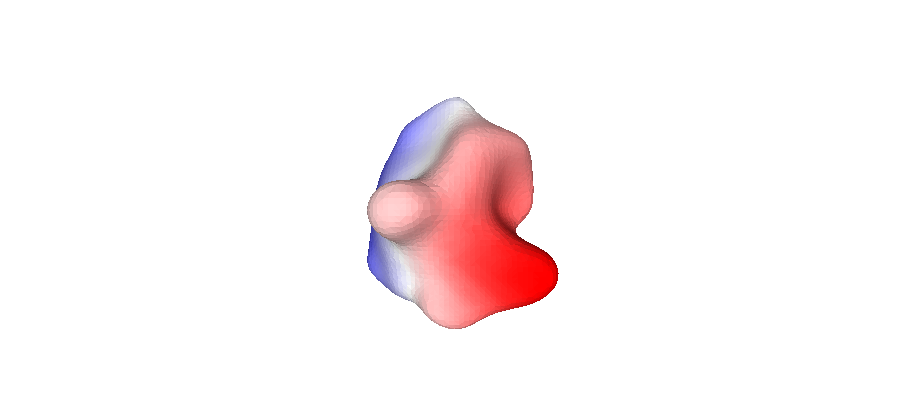}\\[-1mm]$\mathbf{g}^{(N-1)}$}; \&
            \node[rotate=0, align=center,inner sep=0mm] (A4) {\includegraphics[angle=90,width=15mm,trim=270 40 260 60,clip=true]{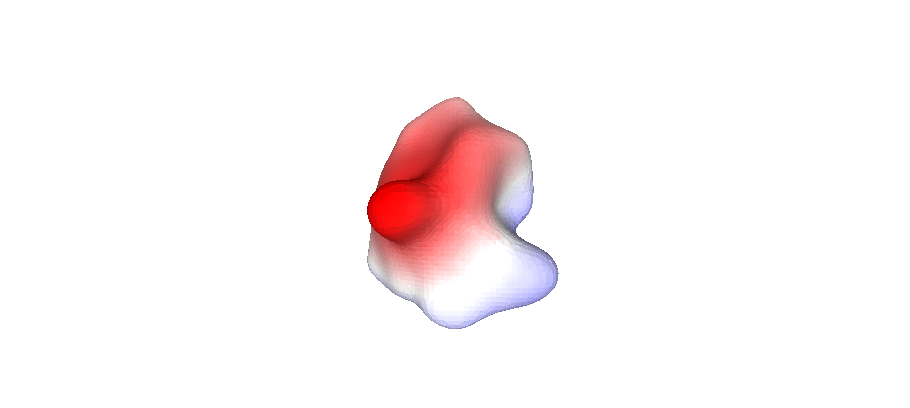}\\[-1mm]$\mathbf{g}^{(N)}$}; \\
            \node[minimum height=2mm] {}; \&
            \node[minimum height=2mm] {}; \&
            \node[minimum height=2mm] {}; \&
            \node[minimum height=2mm] {}; \&
            \node[minimum height=2mm] {}; \&
            \node[minimum height=2mm] {}; \\
            \node[rotate=90, align=center,inner sep=0mm] {\sf\scriptsize Latent\\[-1mm]\sf\scriptsize Layer}; \&
            \node[rotate=0, align=center,inner sep=0mm] (B1) {\includegraphics[angle=90,width=15mm,trim=270 40 260 60,clip=true]{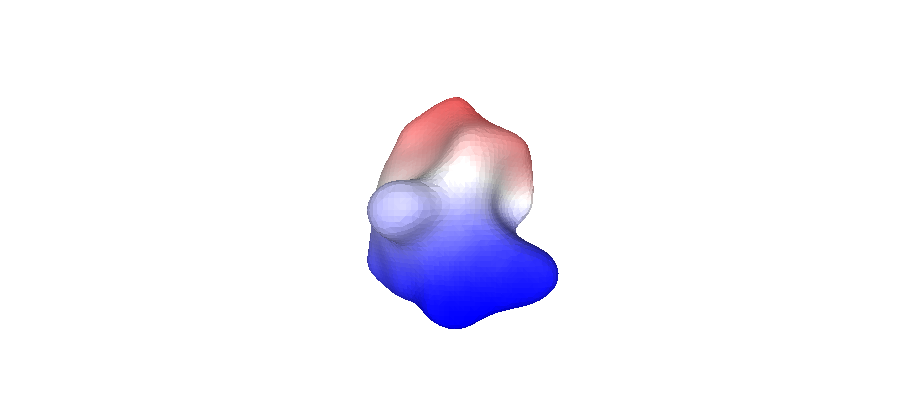}\\[-1mm]$\mathbf{m}^{(1)}$}; \&
            \node[rotate=0, align=center,inner sep=0mm] (B2) {\includegraphics[angle=90,width=15mm,trim=270 40 260 60,clip=true]{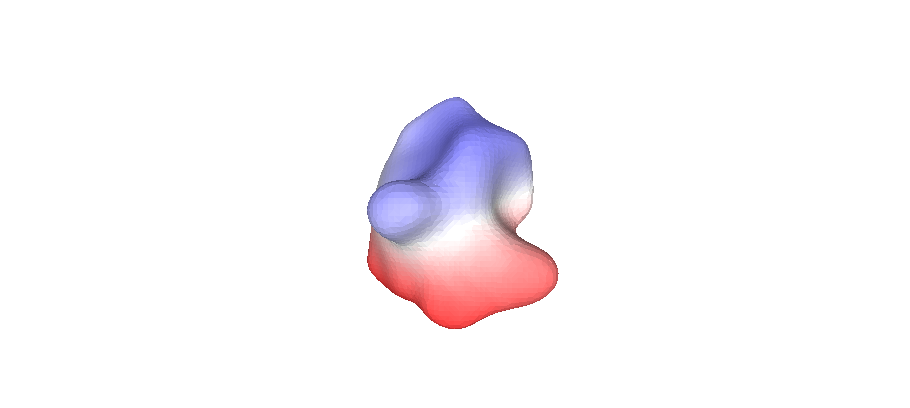}\\[-1mm]$\mathbf{m}^{(2)}$}; \&
            \node[rotate=0, align=center,inner sep=0mm] {\ldots}; \&
            \node[rotate=0, align=center,inner sep=0mm] (B3) {\includegraphics[angle=90,width=15mm,trim=270 40 260 60,clip=true]{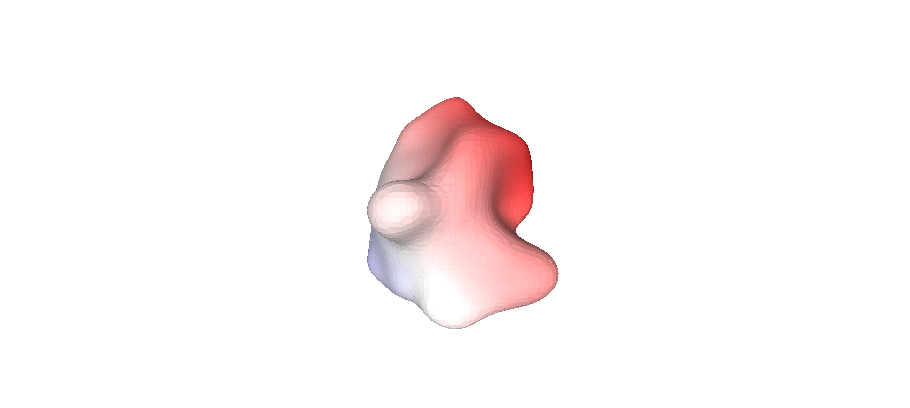}\\[-1mm]$\mathbf{m}^{(K-1)}$}; \&
            \node[rotate=0, align=center,inner sep=0mm] (B4) {\includegraphics[angle=90,width=15mm,trim=270 40 260 60,clip=true]{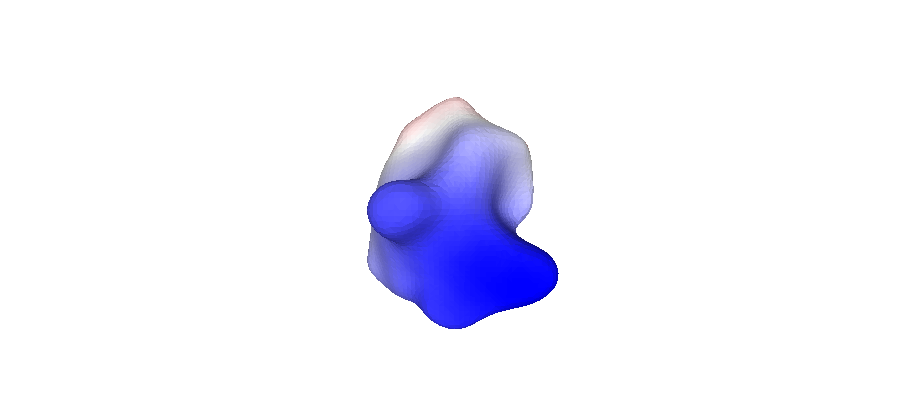}\\[-1mm]$\mathbf{m}^{(K)}$}; \\
            \node[minimum height=2mm] {}; \&
            \node[minimum height=2mm] {}; \&
            \node[minimum height=2mm] {}; \&
            \node[minimum height=2mm] {}; \&
            \node[minimum height=2mm] {}; \&
            \node[minimum height=2mm] {}; \\
            \node[rotate=90, align=center,inner sep=0mm] {\sf\scriptsize Seg.}; \&
            \node[rectangle,text width=10mm,minimum height=10mm, outer sep=0pt] (C1) {}; \&
            \node[rectangle,text width=10mm,minimum height=10mm, outer sep=0pt] (C2) {}; \&
            \node[rectangle,text width=10mm,minimum height=10mm, outer sep=0pt] {}; \&
            \node[rectangle,text width=10mm,minimum height=10mm, outer sep=0pt] (C3) {}; \&
            \node[rectangle,text width=10mm,minimum height=10mm, outer sep=0pt] (C4) {}; \\
        };
        \draw[-latex] (A1.south) -- (B1.north);
        \draw[-latex] ([xshift=1mm]A1.south) -- ([xshift=-1mm]B2.north);
        \draw[-latex] ([xshift=2mm]A1.south) -- ([xshift=-2mm]B3.north);
        \draw[-latex] ([xshift=6mm]A1.south) -- ([xshift=-6mm]B4.north);
        \draw[-latex] ([xshift=-1mm]A2.south) -- ([xshift=1mm]B1.north);
        \draw[-latex] (A2.south) -- (B2.north);
        \draw[-latex] ([xshift=1mm]A2.south) -- ([xshift=-1mm]B3.north);
        \draw[-latex] ([xshift=4mm]A2.south) -- ([xshift=-4mm]B4.north);
        \draw[-latex] ([xshift=-4mm]A3.south) -- ([xshift=4mm]B1.north);
        \draw[-latex] ([xshift=-1mm]A3.south) -- ([xshift=1mm]B2.north);
        \draw[-latex] (A3.south) -- (B3.north);
        \draw[-latex] ([xshift=1mm]A3.south) -- ([xshift=-1mm]B4.north);
        \draw[-latex] ([xshift=-6mm]A4.south) -- ([xshift=6mm]B1.north);
        \draw[-latex] ([xshift=-2mm]A4.south) -- ([xshift=2mm]B2.north);
        \draw[-latex] ([xshift=-1mm]A4.south) -- ([xshift=1mm]B3.north);
        \draw[-latex] (A4.south) -- (B4.north);
6       \node[rectangle,text width=40mm,minimum height=24mm, outer sep=0pt] (D1) [fit=(C1)(C4)] {\\\includegraphics[angle=90,width=15mm,trim=270 40 260 60,clip=true]{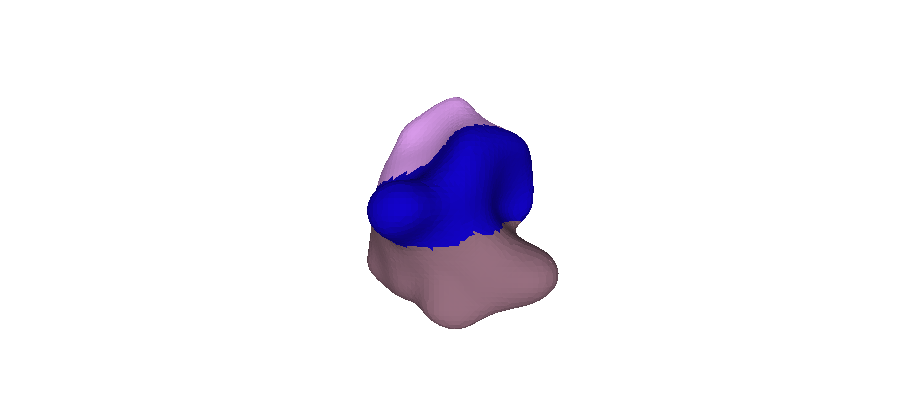}\\[-2mm]$\mathbf{s}$};
        \draw[-latex] (B1.south) -- ([xshift=-10mm,yshift=-8mm]D1.north);
        \draw[-latex] (B2.south) -- ([xshift=-6mm,yshift=-6mm]D1.north);
        \draw[-latex] (B3.south) -- ([xshift=6mm,yshift=-6mm]D1.north);
        \draw[-latex] (B4.south) -- ([xshift=10mm,yshift=-8mm]D1.north);
    \end{tikzpicture}
\vspace{-6mm}
\caption{%
Visualization of geometric bases and spatial segmentation model.
The top row shows the geometric bases as the values vary along the geometric
surface (blue is negative, white is near zero, and red is positive).
The middle row shows the latent layers that are a linear combination of the
geometric bases.
The bottom row shows an example segmentation produced by these latent layers
(each solid color is a distinct material region).
By taking a linear combination of these bases, we can represent a large variety
of segmentations of the geometry surface compactly.
}
\label{fig:eiglaplace}
\end{figure}
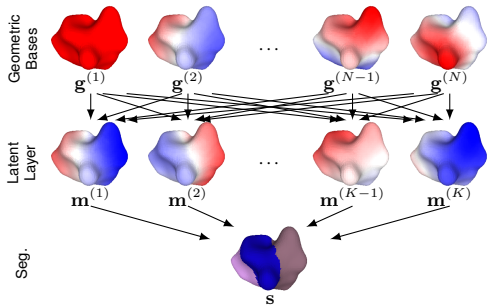

We now develop a spatial segmentation model that can dissect the geometry into $K$ material regions. As is often the case in the real world, we want the segmentation model to prefer contiguous regions. To do this, we construct $K$ latent material layers, $\mathbf{m}$. Each latent material layer consists of a set of scalar values that varies across the surface of the geometry. The values of the latent layers are expressed as a linear combination of
geometric basis functions,
\begin{equation}
\mathbf{m}_{\mathbf{x}}^{(k)} = \sum_{n=1}^{N} \mathbf{g}_{\mathbf{x}}^{(n)} \mathbf{a}_{k}^{(n)},
\end{equation}
where $\mathbf{g}$ are the $N$ geometric basis functions (we use 32 in practice), $\mathbf{a}$ are the coefficients, and $\mathbf{x}$ is a surface point. We determine the material of point $\mathbf{x}$ by taking the layer whose latent material layer is maximum among the $K$ layers,
\begin{equation}
f(\theta_h, \theta_d; \mathbf{x}) = f(\theta_h, \theta_d; \mathbf{R}_{\arg\max_k \mathbf{m}_{\mathbf{x}}^{(k)}}),
\end{equation}
where the coefficients $\mathbf{a}$ dictate the segmentation of the 3D geometry. This approach is advantageous because the spatial segmentation can be modeled compactly as contiguous regions with the appropriate set of geometric bases. Figure \ref{fig:eiglaplace} gives a visual description of the spatial segmentation model.

Choosing the correct set of geometric bases is critical. The bases should vary smoothly to help ensure that there are large contiguous material regions and allow material changes where we might expect two different materials to meet (e.g., in regions of sudden high curvature). The $k$ largest eigenvectors of a matrix of an appropriately formulated similarity matrix between triangles will meet these requirements. We first construct a distance metric between triangles to take a weighted combination of geodesic distance, angle, and color of adjacent triangles. To compute the color of each triangle, we simply average the input pixel values of each triangle and use nearest neighbor to fill in triangles with no image pixels. A full distance matrix is computed using all-pairs shortest path. The similarity matrix is then computed $s_{i,j} = \exp\left[ -d_{i,j}^2/\sigma^2 \right]$. Taking the eigenvectors with the largest corresponding eigenvalues will give us a set of geometric bases that prefers material region boundaries along geometric and photometric seams.

\subsection{Spatially-Varying Reflectance}

We must also handle reflectance variation within material regions. We observe that while one material region may have a relatively constant reflectance behavior along a large surface area, the diffuse texture can change within material regions with high frequency. To address this, we allow the diffuse color---$c_{j',\lambda}$ for the least specular lobe $j'$---to vary along the surface independently of the fundamental material regions. We drop the dependency on color channel ($\lambda$) here for brevity and let $\tilde{\mathbf{c}} = \mathbf{c}_{j'}$ be the diffuse texture which is accessed by a surface point $\mathbf{u}$.
%
%

We must constrain the diffuse texture carefully to allow for enough flexibility while preventing overfitting (e.g., explaining away non-texture appearance such as highlights). We adopt two strategies found in previous work for constraining the diffuse texture. Alldrin et al. showed the albedo distribution of a surface (i.e., diffuse texture) usually has low entropy, which they used to resolve bas-relief ambiguity in photometric stereo. We place an entropy prior on the diffuse texture map,
\begin{equation}
p_1(\tilde{\mathbf{c}}) = \exp\left[ -H(\tilde{\mathbf{c}}) \right],
\end{equation}
where $E$ is the entropy of texture map $\tilde{\mathbf{c}}$. We also apply a spatial smoothness prior on the texture map,
\begin{equation}
p_2(\tilde{\mathbf{c}}) \propto \prod_{\mathbf{u},\mathbf{v}} \exp\left[ - \left(\tilde{\mathbf{c}}_{\mathbf{u}} - \tilde{\mathbf{c}}_{\mathbf{v}} \right)^2  w(d(\mathbf{u}, \mathbf{v}))  \right],
\end{equation}
where $\mathbf{u}, \mathbf{v}$ are surface points, $\tilde{\mathbf{c}}_{\mathbf{u}}$ is the diffuse color of surface point $\mathbf{u}$, and $w$ is a weighting function based on the distance $d(\mathbf{u}, \mathbf{v})$ between surface points $\mathbf{u}$ and $\mathbf{v}$. This spatial smoothness prior enforces similar diffuse values strongly when the surface points are close and decays for surface points that are farther apart.

\section{Representing Scene Illumination}

We model the illumination of a scene with an infinitely-distant spherical panorama which we refer to as the illumination map. This is a good choice for representing natural illumination because it enables modeling of light sources of almost any size, shape, and color. Additionally, it is important to recover a non-parametric representation of the illumination to capture fine details that generic bases like spherical harmonics or wavelets are unable to capture. Much of the priors on the illumination map follow past work with the important exception of a prior on the color to resolve the color constancy problem.

Similar to past work \cite{Lombardi_ECCV12,Lombardi_PAMI15}, we place a sparsity prior on gradients of the illumination map but specifically on its logarithm
\small
\begin{equation}
    p_1(\mathbf{L}) \propto \prod_{\theta, \phi} \exp\left[ -b^{-1} \left( \sqrt{ \sum_{\lambda} \frac{\partial L_{\theta, \phi, \lambda}}{\partial \theta}^2 + \frac{\partial L_{\theta, \phi, \lambda}}{\partial \phi}^2 } \right)^{\alpha} \right],
\end{equation}
\normalsize
as this affords us some robustness to the scale of the input pixel values.

Following past work \cite{Lombardi_ECCV12}, we also place a prior on the entropy of the illumination map to prefer illumination maps with large entropy to resolve the frequency ambiguity between illumination and reflectance by preferring reflectance to explain the frequency attenuation. The entropy prior has the form,
\begin{equation}
p_2(\mathbf{L}) = \exp\left[ -H(\mathbf{L}) \right],
\end{equation}
where $H$ is the Entropy of illumination map $\mathbf{L}$.

In addition to these priors on the intensity, we also introduce a novel chromaticity prior that prefers unsaturated colors in the illumination map with a Dirichlet distribution,
\small
\begin{equation}
p_3(\mathbf{L}) \propto \prod_{\theta, \phi, \lambda} \left( \frac{L_{\theta, \phi, \lambda}}{\sum_{\lambda'} L_{\theta, \phi, \lambda'}} \right)^{\alpha - 1}.
\end{equation}
\normalsize
This prior helps enforce the grey-world assumption, that the illumination environment is, on average, uncolored so that the color constancy ambiguity can be resolved.

\section{Modeling Indirect Illumination}

A major problem for reflectance and illumination estimation in the real world is indirect illumination (i.e., light that arrives at a surface from other non-emitting surfaces). Methods that ignore indirect illumination will incorrectly attribute interreflection or shadows with changes in the reflectance, illumination, or geometry of a surface, causing important information about the scene to be lost. 

To handle indirect illumination during the evaluation of the posterior, we compute the predicted radiance $\hat{\mathbf{I}}(\mathbf{p}; \mathbf{R}, \mathbf{L}, \mathbf{G})$ at pixel $\mathbf{p}$ with an unbiased rendering algorithm.
%
Formally, we write the predicted pixel value,
\begin{equation}
\hat{\mathbf{I}}_{\lambda}(\mathbf{p}; \mathbf{R}, \mathbf{L}, \mathbf{G}) =
    \int_{\mathbf{v}} E_{\lambda}(t(\mathbf{e}, \mathbf{v}; \mathbf{G}), -\omega_i) d\mathbf{v},\label{eq:pixelirradiance}
\end{equation}
where $t(\mathbf{e}, \mathbf{v}; \mathbf{G})$ is a function that computes the closest intersection point on the surface of the geometry $\mathbf{G}$ with the
ray from eye position $\mathbf{e}$ along viewing direction $\mathbf{v}$, $\mathbf{v}$ is a set of directions passing through pixel $\mathbf{p}$, and $E_{\lambda}(\mathbf{x}, \omega_o)$ is the radiance coming from surface point $\mathbf{x}$ to the camera. The radiance from surface point $\mathbf{x}$ in direction $\omega_o$, $E_{\lambda}(\mathbf{x}, \omega_o)$, is written,
\small
\begin{equation}
E_{\lambda}(\mathbf{x}, \mathbf{\omega}_o) =
D(\mathbf{x}, \omega_o) +
\int_{\Omega} f_{\lambda}(\mathbf{x}, \omega_i, \mathbf{\omega}_o) E_{\lambda}(t(\mathbf{x},\mathbf{\omega}_i; \mathbf{G}), -\omega_i) \bar{v}(\mathbf{x}, \omega_i) d\omega_i,
\end{equation}
\normalsize
where
\small
\begin{equation}
D(\mathbf{x}, \omega_o) =
\int_{\Omega} f_{\lambda}(\mathbf{x}, \omega_i, \mathbf{\omega}_o) \mathbf{L}_i(\omega_i) v(\mathbf{x}, \omega_i) d\omega_i,
\end{equation}
\normalsize
is the direct lighting integral, $v$ is a visibility function that determines if ray beginning at $\mathbf{x}$ in direction $\omega_i$ intersects the geometry, and
$f_{\lambda}(\mathbf{x}, \omega_i, \omega_o)$ is the BRDF at surface point
$\mathbf{x}$.
This is a rewriting of the rendering equation \cite{Kajiya_SG86} into a direct and indirect lighting component. We solve this equation using Monte Carlo integration.

\section{Refining Rough Geometry}

To best understand how to refine the geometry, we must consider the characteristics of the initial RGB-D geometry. There are two main sources of error when using a series of registered and combined RGB-D images: details finer than the resolution of the depth images and large-scale low-frequency errors that arise while fusing the data into a single geometric model. These sources of error suggest a targeted approach for refining the input geometry. We attempt to correct the large-scale low-frequency error by maximizing the posterior while modeling the mesh vertices using a set of smoothly varying bases. We can reuse the geometric bases we use for the scene segmentation for this purpose.

We write the vertices as an initial vertex $\mathbf{V}^{(0)}$ plus a linear combination of the geometric bases in the direction of the surface normal at that point,
\begin{equation}
\mathbf{V}_{\mathbf{x}} = \mathbf{V}_{\mathbf{x}}^{(0)} + \sum_{j=1}^{J} w_j g_{\mathbf{x}}^{(j)} \mathbf{N}_{\mathbf{x}},
\end{equation}
where $\mathbf{x}$ is a point on the surface, $\mathbf{w}$ is a set of coefficients that controls the influence of the geometric bases, and $\mathbf{N}_{\mathbf{x}}$ is the surface normal at point $\mathbf{x}$. This representation allows the geometry to slightly ``inflate'' or ``deflate'' in a spatially coherent way to best match the input images. After optimizing large-scale variation, we recompute the surface normals of each
vertex by averaging the neighboring triangle surface normals.

\section{MAP Estimation}

The radiometric scene decomposition is achieved by maximizing the posterior distribution using gradient-based optimization. For this, we derive an algorithm analogous to path tracing that computes the gradients of the posterior with respect to the reflectance and illumination so that an efficient implementation of path tracing rendering (e.g., using GPU) can readily be used for the MAP estimation. We also carefully initialize the variables and adopt an ordered introduction of the variables into the optimization to avoid unwanted local minima.

%
%


\subsection{Likelihood Gradients}

First, we rewrite the radiance $E$ with an expression that easily translates into an iterative path tracing algorithm. In a path tracing algorithm, a light path is traced from the camera position outward to the geometry. Each time the light ray intersects the geometry, the direct lighting integral is computed at that point and then a new ray direction is selected to continue the light path which terminates when a light source is struck. We rewrite $E$,
\begin{equation}
E(\mathbf{x}, \mathbf{\omega}_o) =
\sum_{i=0}^{\infty} \prod_{j}^{i} f\left(\mathbf{x}^{(j)}, \mathbf{\omega}_i^{(j)}, \mathbf{\omega}_o^{(j)}\right)
%
D(\mathbf{x}^{(j)}, \omega_o^{(j)}),
\end{equation}
where
$\omega_i^{(j)}, \omega_o^{(j)}, \mathbf{x}^{(j)}$ are the incident, exitant, and intersection point of the $j^{\mathrm{th}}$ point on the light path. This better represents the steps in an iterative path tracing algorithm: we first trace a ray to the surface, keeping track of an attenuation parameter which is the product of all BRDFs seen so far, and then we solve for the direct lighting component which is the integral, before continuing the path tracing by shooting a ray to the next surface. This form makes it easier to leverage an iterative path tracing-like algorithm to compute the gradients.

We can now write the derivative of the radiance $E$,
%
\small
\begin{equation}
\frac{\partial E(\mathbf{x}, \mathbf{\omega}_o)}{\partial \mathbf{R}} =
\sum_{i=0}^{\infty} \left(
\frac{ \partial \prod_{j}^{i} f\left(\mathbf{x}^{(j)}, \mathbf{\omega}_i^{(j)}, \mathbf{\omega}_o^{(j)}\right) }{ \partial\mathbf{R}}
%
D(\mathbf{x}, \omega_o) +
\prod_{j}^{i} f\left(\mathbf{x}^{(j)}, \mathbf{\omega}_i^{(j)}, \mathbf{\omega}_o^{(j)}\right)
%
\frac{\partial D(\mathbf{x}, \omega_o)}{\partial\mathbf{R}} \right).
\end{equation}
\normalsize
We write the derivative of the product of the BRDFs as a product of two factors,
\small
\begin{equation}
\frac{ \partial \prod_{j}^{i} f\left(\mathbf{x}^{(j)}, \mathbf{\omega}_i^{(j)}, \mathbf{\omega}_o^{(j)}\right) }{ \partial\mathbf{R}} =
\left( \prod_{j}^{i} f\left(\mathbf{x}^{(j)}, \mathbf{\omega}_i^{(j)}, \mathbf{\omega}_o^{(j)}\right) \right)
\left( \sum_{j}^{i} \frac{\frac{\partial f\left(\mathbf{x}^{(j)}, \mathbf{\omega}_i^{(j)}, \mathbf{\omega}_o^{(j)}\right)}{\partial\mathbf{R}}}{f\left(\mathbf{x}^{(j)}, \mathbf{\omega}_i^{(j)}, \mathbf{\omega}_o^{(j)}\right)} \right).
\end{equation}
\normalsize
This allows us to easily see how to compute the derivative in a path tracing algorithm: we can iteratively compute the new sum term just like the attenuation factor.

\begin{figure}[t]
\vspace{1mm}
\hrule
\vspace{1mm}
\textbf{Algorithm 1:} Compute Predicted Pixel Irradiance Gradient with Respect to Reflectance
\vspace{1mm}
\hrule
\begin{algorithmic}
\State $A[1 \times 3] \leftarrow 1$ \Comment{attenuation factor}
\State $\partial A/\partial \mathbf{R}[|\mathbf{R}| \times 3] \leftarrow 0$

\State $E[1 \times 3] \leftarrow 0$ \Comment{outgoing radiance}
\State $\partial E/\partial \mathbf{R}[|\mathbf{R}| \times 3] \leftarrow 0$

\State $r$ is a ray from the camera location to the pixel

\While{$r$ intersects the geometry}
    \State $\mathbf{x}$ is the intersection point of ray $r$ with the scene


    \State $D[1 \times 3] \leftarrow 0$ \Comment{direct lighting integral}
    \State $\partial D/\partial \mathbf{R}[|\mathbf{R}| \times 3] \leftarrow 0$

    \For{$1 \ldots $ N} \Comment{compute direct lighting}
        \State $s \leftarrow $ random ray direction
        \If{ray $s$ hits the illumination environment}
            \State $D \leftarrow D + f(r, s; \mathbf{x}) * \mathbf{L}(s) / p(s)$
            \State $\partial D/\partial \mathbf{R} \leftarrow \partial D/\partial \mathbf{R} + \frac{\partial f(r, s; \mathbf{x})}{\partial \mathbf{R}} * \mathbf{L}(s) / p(s)$
        \EndIf
    \EndFor

    \State $E \leftarrow E + A * D$
    \State $\partial E/\partial \mathbf{R} \leftarrow \partial E/\partial \mathbf{R} + A * (\partial A/\partial \mathbf{R} * D + \partial D/\partial \mathbf{R})$

    \State $r' \leftarrow $ random ray direction

    \State $A \leftarrow A * f(r', r; \mathbf{x})$
    \State $\partial A/\partial \mathbf{R} \leftarrow \partial A/\partial \mathbf{R} + \frac{ \partial f(r', r; \mathbf{x})/\partial \mathbf{R} }{ f(r', r; \mathbf{x}) }$

    \State $r \leftarrow r'$

\EndWhile
\end{algorithmic}
\vspace{1mm}
\hrule
\vspace{-6mm}
\label{fig:algo:ptgrad}
\end{figure}

Algorithm 1 gives the complete steps of an algorithm to efficiently compute the gradient of the radiance $E$ with respect to the reflectance parameters that handles indirect illumination. Note that it has the same asymptotic complexity as path tracing. We can use a similar algorithm for taking the derivative with respect to the natural illumination and surface normals.


Numerically computing the gradient requires Monte Carlo integration of the direct lighting integral and the gradient of the direct lighting integral. Multiple importance sampling is typically used to compute the direct lighting integral by separately sampling from a BRDF proposal distribution and an illumination proposal distribution and then appropriately weighting the samples. We use the same technique for computing the gradient of the direct lighting integral. When computing the gradient of the direct lighting integral with respect to the BRDF, we use a standard BRDF proposal distribution. When computing the gradient of the direct lighting integral with respect to the illumination, the illumination proposal distribution only samples rays in the direction of the illumination map pixel $L_{\theta,\phi}$ whose gradient is being taken.

\subsection{Initialization and Ordered Optimization}

Directly minimizing the negative log posterior to compute a MAP estimate is difficult because of the complex objective function. To deal with this, we propose a careful initialization of reflectance and illumination. We initialize our illumination environment by first estimating a uniform gray-scale illumination map to estimate the general scale of the intensity values of the scene. The reflectance coefficients $\mathbf{\Psi}$ are initialized to zero, which corresponds to an average BRDF in the DSBRDF model. Before estimating the reflectance coefficients, we estimate the lobe diffuse colors $\mathbf{c}$ which helps prevent the illumination map from overfitting by trying to represent the colors of the objects in the scene. The scene segmentation is initialized by performing a clustering of the triangle mesh using the geometric bases $\mathbf{g}$ and the color of each triangle.

We found that facilitating the order of optimization is important to avoid local minima. First, we alternate between estimating the reflectance and illumination until convergence. After that has converged, we incorporate the segmentation estimation, then we incorporate diffuse texture estimation, and finally we incorporate geometry estimation. This procedure allows for a sort of ``coarse-to-fine'' estimation scheme, where we begin by estimating broad areas of reflectance in the scene, and later begin to fill in the fine detail such as the diffuse texture. Convergence is guaranteed as each sub-optimization only minimizes the error. Although no guarantees on reaching global minimum can be given as clearly the error function is not convex, we empirically find the initialization and ordered optimization to avoid the scene decomposition from getting trapped in unwanted local minima. Reflectance and illumination estimation typically finishes within a few hours whereas diffuse texture estimation can take up to a day or two depending on the resolution of the texture map using a PC with a single GPU.

\section{Experimental Results}\label{sec:exp}

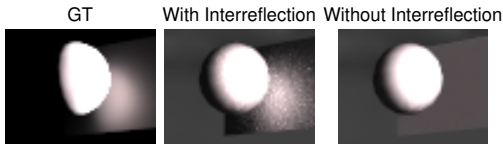
\begin{figure}[t]
\centering
    \begin{tikzpicture}[inner sep=0mm]
        \matrix (table) [row sep=1mm, column sep=1mm, ampersand replacement=\&] {
            \node[rotate=0, inner sep=0mm] {\sf\scriptsize GT}; \&
            \node[rotate=0, inner sep=0mm] {\sf\scriptsize With Interreflection}; \&
            \node[rotate=0, inner sep=0mm] {\sf\scriptsize Without Interreflection}; \\
            \node {\includegraphics[width=20mm,trim=0 4 0 8,clip=true]{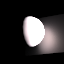}}; \&
            \node {\includegraphics[width=20mm,trim=0 4 0 8,clip=true]{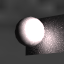}}; \&
            \node {\includegraphics[width=20mm,trim=0 4 0 8,clip=true]{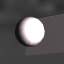}}; \\
        };
    \end{tikzpicture}
\caption{%
The importance of considering indirect illumination.
%
Left: An image of a simple scene consisting of a sphere and a quad.
%
%
Middle: Our method models interreflection and correctly estimates the reflectance of the quad.
Right: When interreflection is not modeled, the algorithm cannot correctly estimate the reflectance of this scene.
}
\label{fig:results:interreflecttest}
\end{figure}

\begin{figure}[t!]
\centering
    \begin{tikzpicture}[inner sep=0mm]
        \matrix (table) [row sep=1mm, column sep=1mm, ampersand replacement=\&] {
            \node[rotate=0, inner sep=0mm] {}; \&
            \node[rotate=0, inner sep=0mm] {\sf\scriptsize Best}; \&
            \node[rotate=0, inner sep=0mm] {\sf\scriptsize Best}; \&
            \node[rotate=0, inner sep=0mm] {\sf\scriptsize Typical}; \&
            \node[rotate=0, inner sep=0mm] {\sf\scriptsize Typical}; \&
            \node[rotate=0, inner sep=0mm] {\sf\scriptsize Worst}; \\
            \node[rotate=90, align=center,inner sep=0mm] {\sf\scriptsize Estimated\\[-2mm]\sf\scriptsize Appearance}; \&
            \node[rotate=0, inner sep=0mm] {\includegraphics[width=15mm]{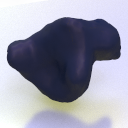}}; \&
            \node[rotate=0, inner sep=0mm] {\includegraphics[width=15mm]{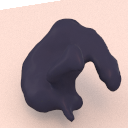}}; \&
            \node[rotate=0, inner sep=0mm] {\includegraphics[width=15mm]{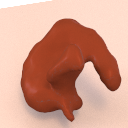}}; \&
            \node[rotate=0, inner sep=0mm] {\includegraphics[width=15mm]{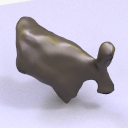}}; \&
            \node[rotate=0, inner sep=0mm] {\includegraphics[width=15mm]{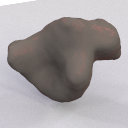}}; \\
            \node[rotate=90, align=center,inner sep=0mm] {\sf\scriptsize GT\\[-2mm]\sf\scriptsize Appearance}; \&
            \node[rotate=0, inner sep=0mm] {\includegraphics[width=15mm]{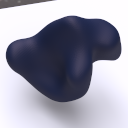}}; \&
            \node[rotate=0, inner sep=0mm] {\includegraphics[width=15mm]{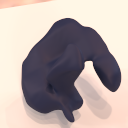}}; \&
            \node[rotate=0, inner sep=0mm] {\includegraphics[width=15mm]{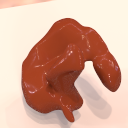}}; \&
            \node[rotate=0, inner sep=0mm] {\includegraphics[width=15mm]{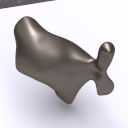}}; \&
            \node[rotate=0, inner sep=0mm] {\includegraphics[width=15mm]{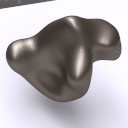}}; \\
            %
            \node[rotate=90, align=center,inner sep=0mm] {\sf\scriptsize Est.\\[-2mm]\sf\scriptsize Illum.}; \&
            \node[rotate=0, inner sep=0mm] {\includegraphics[width=15mm]{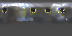}}; \&
            \node[rotate=0, inner sep=0mm] {\includegraphics[width=15mm]{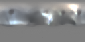}}; \&
            \node[rotate=0, inner sep=0mm] {\includegraphics[width=15mm]{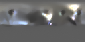}}; \&
            \node[rotate=0, inner sep=0mm] {\includegraphics[width=15mm]{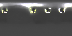}}; \&
            \node[rotate=0, inner sep=0mm] {\includegraphics[width=15mm]{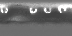}}; \\
            \node[rotate=90, align=center,inner sep=0mm] {\sf\scriptsize GT\\[-2mm]\sf\scriptsize Illum.}; \&
            \node[rotate=0, inner sep=0mm] {\includegraphics[width=15mm]{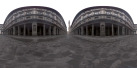}}; \&
            \node[rotate=0, inner sep=0mm] {\includegraphics[width=15mm]{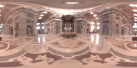}}; \&
            \node[rotate=0, inner sep=0mm] {\includegraphics[width=15mm]{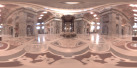}}; \&
            \node[rotate=0, inner sep=0mm] {\includegraphics[width=15mm]{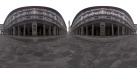}}; \&
            \node[rotate=0, inner sep=0mm] {\includegraphics[width=15mm]{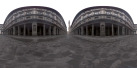}}; \\
        };
    \end{tikzpicture}
\caption{
Examples from quantitative evaluation on an extensive set of synthetic images.
%
%
%
%
The first row shows a novel view of several scenes rendered with the recovered
reflectance, illumination, and geometry compared to the ground truth in the second row.
The third row shows our estimated illumination map compared to the ground truth
illumination in the fourth row.
}
\label{fig:results:geomtest}
\end{figure}
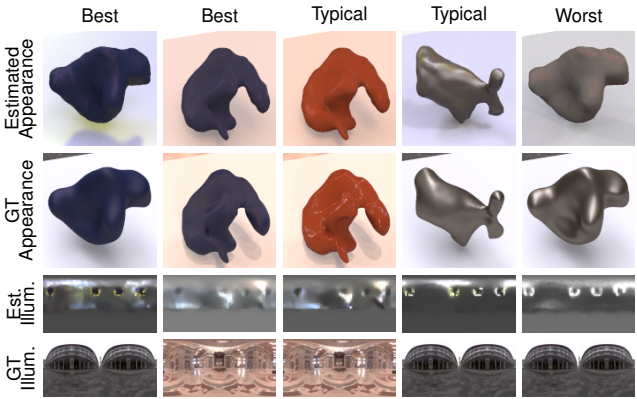

\begin{figure*}[t]
\centering
\captionsetup[subfigure]{labelformat=empty}
\subfloat[]{
    \begin{minipage}{0.6\linewidth}
    \begin{tikzpicture}[inner sep=0mm]
        \matrix (table) [row sep=1mm, column sep=1mm, ampersand replacement=\&] {
            \node[rotate=0, inner sep=0mm] {}; \&
            \node[rotate=0, inner sep=0mm] {\scriptsize \textsf{Estimated}}; \&
            \node[rotate=0, inner sep=0mm] {\scriptsize \textsf{Input (1 of 3)}}; \&
            \node[rotate=0, inner sep=0mm] {}; \&
            \node[rotate=0, inner sep=0mm] {\scriptsize \textsf{Estimated}}; \&
            \node[rotate=0, inner sep=0mm] {\scriptsize \textsf{Ground Truth}}; \\
            \node[rotate=90, inner sep=0mm] {\scriptsize \textsf{Appearance}}; \&
            \node {\includegraphics[width=11mm]{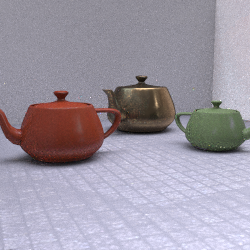}}; \&
            \node {\includegraphics[width=11mm]{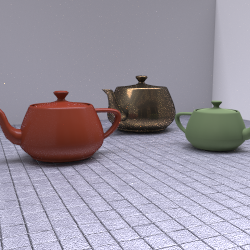}}; \&
            \node[rotate=90, align=center,inner sep=0mm] {\sf\scriptsize Appearance\\[-2mm]\sf\scriptsize (Novel View)}; \&
            \node {\includegraphics[width=11mm]{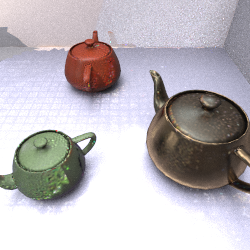}}; \&
            \node {\includegraphics[width=11mm]{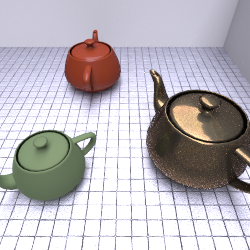}}; \\
            %

            \node[rotate=0, inner sep=0mm] {}; \&
            \node[rotate=0, inner sep=0mm] {\scriptsize \textsf{Estimated}}; \&
            \node[rotate=0, inner sep=0mm] {\scriptsize \textsf{Ground Truth}}; \&
            \node[rotate=0, inner sep=0mm] {}; \&
            \node[rotate=0, inner sep=0mm] {\scriptsize \textsf{Estimated}}; \&
            \node[rotate=0, inner sep=0mm] {\scriptsize \textsf{Ground Truth}}; \\
            \node[rotate=90, inner sep=0mm] {\sf\scriptsize Segmentation}; \&
            \node {\includegraphics[width=15mm,trim=0 40 0 30,clip=true]{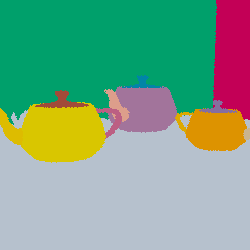}}; \&
            \node {\includegraphics[width=15mm,trim=0 40 0 30,clip=true]{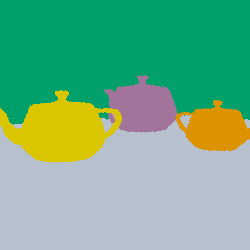}}; \&
            \node[rotate=90, inner sep=0mm] {\scriptsize \textsf{Illumination}}; \&
            \node {\includegraphics[width=15mm]{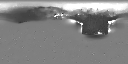}}; \&
            \node {\includegraphics[width=15mm]{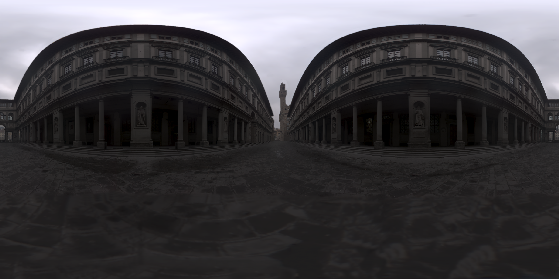}}; \\
        };
    \end{tikzpicture}

    \end{minipage}
}~
\subfloat[]{
    \begin{minipage}{0.4\linewidth}
    \begin{tikzpicture}[inner sep=0mm]
        \matrix (table) [row sep=1mm, column sep=1mm, ampersand replacement=\&] {
            \node [rotate=0, align=center,inner sep=0mm] {}; \&
            \node [rotate=0, align=center,inner sep=0mm] {\sf\scriptsize Estimated\\[-2mm]\sf\scriptsize Reflectance}; \&
            \node [rotate=0, align=center,inner sep=0mm] {\sf\scriptsize Ground Truth\\[-2mm]\sf\scriptsize Reflectance}; \\
            \node [rotate=90, align=center,inner sep=0mm] {\sf\scriptsize Red\\[-2mm]\sf\scriptsize Teapot}; \&
            \node {\includegraphics[width=20mm,trim=0 0 300 0,clip=true]{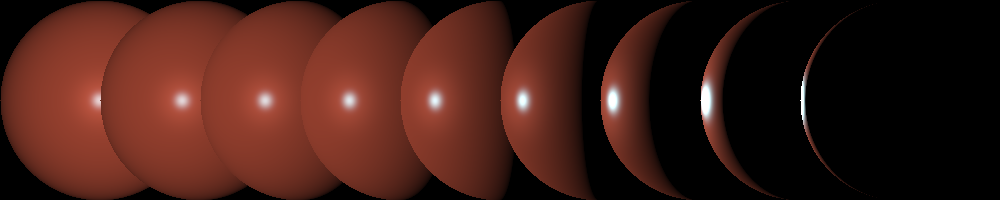}}; \&
            \node {\includegraphics[width=20mm,trim=0 0 300 0,clip=true]{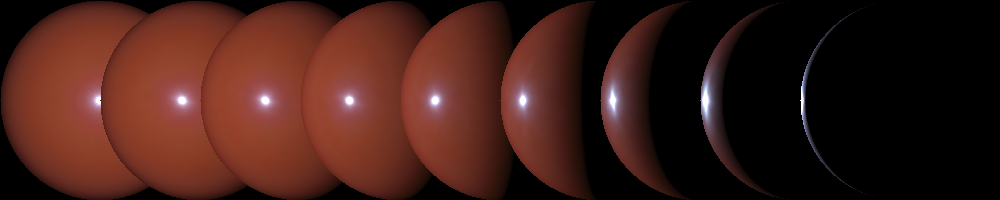}}; \\
            \node [rotate=90, align=center,inner sep=0mm] {\sf\scriptsize Metallic\\[-2mm]\sf\scriptsize Teapot}; \&
            \node {\includegraphics[width=20mm,trim=0 0 300 0,clip=true]{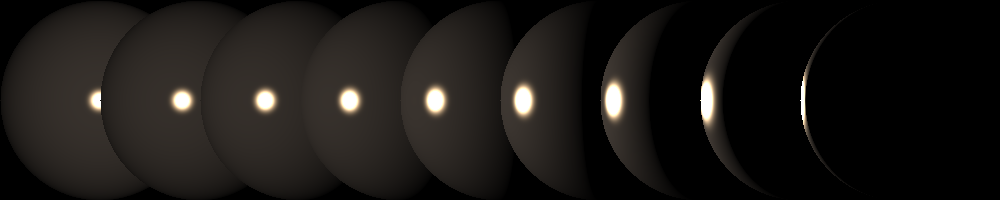}}; \&
            \node {\includegraphics[width=20mm,trim=0 0 300 0,clip=true]{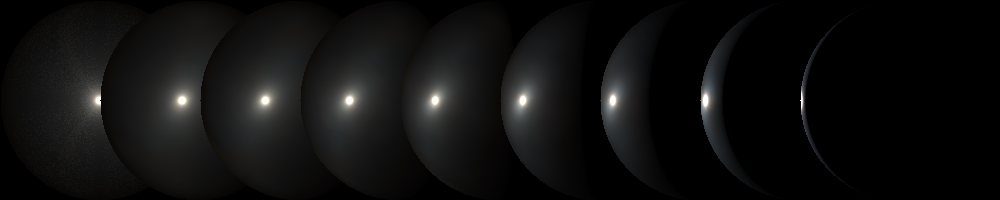}}; \\
            \node [rotate=90, align=center,inner sep=0mm] {\sf\scriptsize Green\\[-2mm]\sf\scriptsize Teapot}; \&
            \node {\includegraphics[width=20mm,trim=0 0 300 0,clip=true]{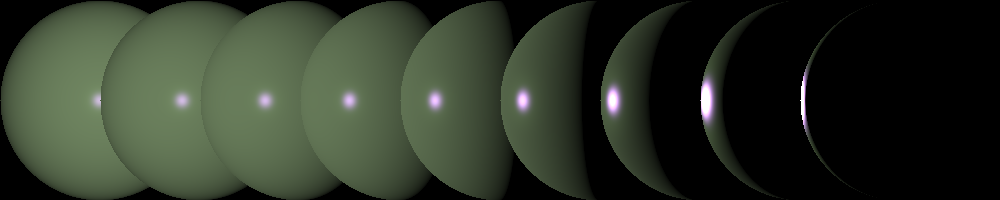}}; \&
            \node {\includegraphics[width=20mm,trim=0 0 300 0,clip=true]{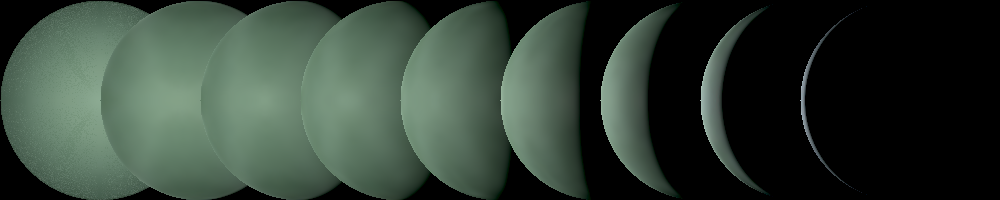}}; \\
        };
    \end{tikzpicture}
    \end{minipage}
}
\vspace{-6mm}
\caption{%
%
Radiometric decomposition results of synthetic scenes.
Top row: predicted image with recovered reflectance and illumination, one of
three input images, predicted image from a novel view of the scene (i.e., not
input to the algorithm), ground truth of novel view.
Bottom row: estimated segmentation map, ground truth segmentation map, estimated
illumination, ground truth illumination.
On the right are cascaded renderings of a sphere as a point light direction is
slowly varied for the recovered and ground truth reflectance for the teapots.
}
\label{fig:results:synth1}
\end{figure*}



We first show the results of a series of experiments using synethic images designed to evaluate different aspects of our method. In each case, the scenes are rendered from a handful of views using PBRT \cite{Pharr_2010} and used as input to our method.

Figure \ref{fig:results:interreflecttest} demonstrates the importance of considering indirect illumination with a scene containing a specular quad reflecting light from an adjacent sphere. Our method accounts for indirect illumination and therefore correctly estimates the reflectance of the quad. When the indirect illumination computation is disabled, the quad is estimated as being diffuse instead.

Figure \ref{fig:results:geomtest} highlights results from an extensive set of experiments using a set of three different meshes rendered with three different MERL BRDFs under three different illumination environments (27 total scenes) from four viewpoints. In these experiments, we artificially perturb the geometry of each mesh by adding noise and then smoothing the vertex positions to evalute the geometry refinement accuracy in addition to the accuracy of the reflecntance and illumination estimation. The figure shows the best, typical, and worst results in this dataset by presenting a novel view of the scene rendered with the recovered reflectance, illumination, and geometry compared to the ground truth. As shown, the method typically estimates reflectance and illumination very close to the ground truth, despite the noisy geometry. It understandably has difficulty with highly specular materials as the appearance of the object is very sensitive to the noisy surface normals. We quantitatively evaluated the reflectance by computing the log-space RMS Error of the recovered reflectance compared to ground truth and found the average error to be 1.97. To quantitatively evaluate the geometry, we compute the distance between each vertex in the estimated mesh to the closest point on the ground-truth geometry and average all distances. In our 27 experiments, the geometry error after using our geometry refinement method was improved, on average, by 25\% compared to the artificially-perturbed initial geometry. Please see the supplementary material for more results\footnote{\url{http://www.cs.drexel.edu/~kon/mvrani_arxiv_sup.pdf}}.
%
%
%
%

Figure \ref{fig:results:synth1} shows results on an example synthetic scene. This particular scene consists of several teapots in a room with a textured floor. This tests the ability of our algorithm to estimate reflectance and illumination from a relatively simple scene. As shown, our method is able to recover accurate reflectance functions close to the ground truth, even for the brass teapot that is reflecting many other objects in the scene. This would be difficult, if not impossible, for methods that don't explicitly account for indirect illumination. The recovered illumination environment captures the general position and intensity of the sky but lacks detail because of the walls blocking most of the environment. The segmentation map is in general accurate but has some small inaccuracies where it represents some of the teapots with multiple distinct but similar reflectance functions. The patterned texture of the floor is recovered but faint in this example due to the smoothing prior we impose to prevent overfitting with the diffuse texture. Please see the supplementary material for more results.




We have collected a novel dataset of real world scenes to evaluate our method. We capture each scene using a Canon EOS 6D camera attached to a Kinect sensor and acquire high dynamic-range (HDR) images of the scene from several different viewing directions. We use the KinectFusion algorithm \cite{Izadi_UIST11,Newcombe_ISMAR11} to combine multiple depth images into a triangle mesh that is used as the input geometry. We also capture the ground truth illumination using a mirror ball placed in the center of the scene.

\begin{figure*}[t!]
\centering
    \begin{tikzpicture}[inner sep=0mm]
        \matrix (table) [row sep=1mm, column sep=1mm, ampersand replacement=\&] {
            \node[rotate=90, align=center,inner sep=0mm] {}; \&
            \node[rotate=0, align=center,inner sep=0mm] {\sf\scriptsize Predicted\\[-2mm]\sf\scriptsize Appearance}; \&
            \node[rotate=0, inner sep=0mm] {\scriptsize \textsf{Input (1 of 3)}}; \&
            %
            \node[rotate=0, align=center,inner sep=0mm] {\sf\scriptsize Predicted\\[-2mm]\sf\scriptsize Appearance\\[-2mm]\sf\scriptsize (Novel View)}; \&
            \node[rotate=0, align=center,inner sep=0mm] {\sf\scriptsize Ground Truth\\[-2mm]\sf\scriptsize Appearance\\[-2mm]\sf\scriptsize (Novel View)}; \&
            \node[rotate=0, align=center,inner sep=0mm] {\sf\scriptsize Estimated\\[-2mm]\sf\scriptsize Segmentation}; \&
            \node[rotate=0, align=center,inner sep=0mm] {\sf\scriptsize Estimated\\[-2mm]\sf\scriptsize Illumination}; \&
            \node[rotate=0, align=center,inner sep=0mm] {\sf\scriptsize Ground Truth\\[-2mm]\sf\scriptsize Illumination}; \\

            %
            \node[rotate=90, align=center,inner sep=0mm] {\sf\scriptsize Scene 1}; \&
            \node {\includegraphics[width=15mm]{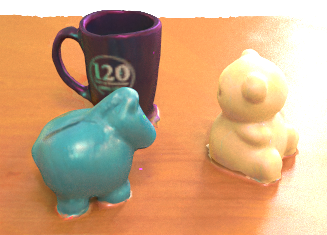}}; \&
            \node {\includegraphics[width=15mm]{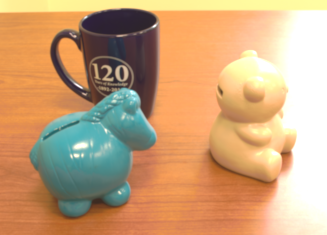}}; \&
            %
            \node {\includegraphics[width=15mm]{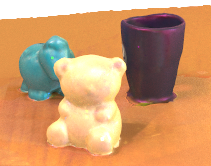}}; \&
            \node {\includegraphics[width=15mm]{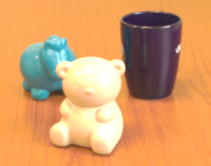}}; \&
            \node {\includegraphics[width=15mm]{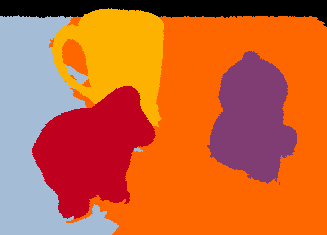}}; \&
            \node {\includegraphics[width=17mm]{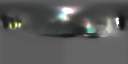}}; \&
            \node {\includegraphics[width=17mm]{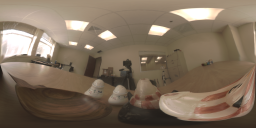}}; \\

            \node[rotate=90, align=center,inner sep=0mm] {\sf\scriptsize Scene 2}; \&
            \node {\includegraphics[width=15mm]{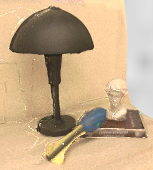}}; \&
            \node {\includegraphics[width=15mm]{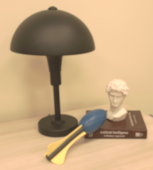}}; \&
            \node {\includegraphics[width=11mm,trim=0 0 0 10,clip=true]{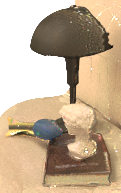}}; \&
            \node {\includegraphics[width=11mm,trim=0 0 0 10,clip=true]{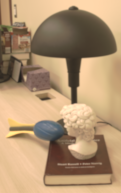}}; \&
            \node {\includegraphics[width=15mm]{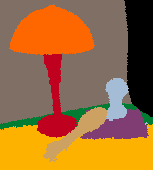}}; \&
            \node {\includegraphics[width=17mm]{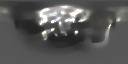}}; \&
            \node {\includegraphics[width=17mm]{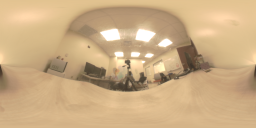}}; \\
        };
    \end{tikzpicture}
\caption{%
Radiometric decomposition results of real-world scenes.
The columns are, from left to right, the predicted appearance of the scene from the
view of the first input image, the first input image of the set, the predicted appearance
from a novel view of the scene, the ground truth image from that novel view,
the estimated segmentation map, the estimated illumination, and the ground truth illumination.
Our method is able to very accurately predict the appearance of the scene from novel views.
}
\label{fig:results:real1}
\end{figure*}
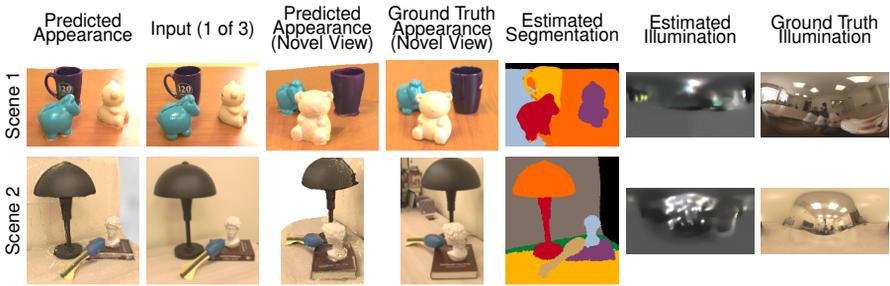

Figure \ref{fig:results:real1} shows results on two real scenes. Note that the first scene would be challenging if the reflectance was assumed to be Lambertian as the light reflecting off the table opposite the viewer would be estimated erroneously as a change in diffuse texture. Our framework is able to correctly deduce that this is due to specular reflectance while also correctly estimating the texture of the table. In this scene we are also able to recover the texture detail on both the table and the coffee mug.

The second scene contains the most geometrically complex arrangement of objects in our dataset. Our method correctly estimates the illumination as being a cluster of lights located behind the viewer although finer details are not captured due to the lack of highly-specular materials in the scene. We also recover plausible reflectance, such as the dull specular highlight on the black lamp, which allows us to accurately predict the appearance of the scene from a novel view. Fine texture details, such as the text on the book, is recovered but with some noise in them. This is mostly in spots where the surface is viewed by few or no input images, and may also be due to small errors in the intrinsic or extrinsic camera parameters. Please see the supplemental material for more results.

Figure \ref{fig:results:stages} show several of the distinct stages of our method. Initially, only reflectance, illumination, and segmentation are estimated. After convergence, texture estimation is incorporated, followed by geometry estimation. This setup helps prevent overfitting (e.g., estimating texture first can cause the texture to model specular highlights). In the figure we see that each step adds important detail: reflectance and illumination captures the specularity of the table, the texture estimation captures the text on the mug, and geometry estimation improves the shape of the mug handle. It clearly shows that all the factors we consider in the formulation are essential for successful radiometric scene decomposition.

\section{Conclusion}

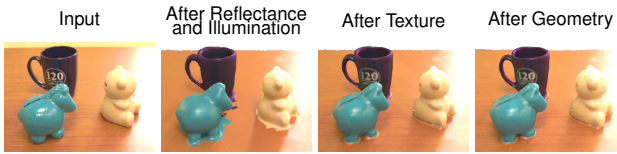
\begin{figure}[t]
\centering
    \begin{tikzpicture}[inner sep=0mm]
        \matrix (table) [row sep=1mm, column sep=1mm, ampersand replacement=\&] {
            \node[rotate=0, align=center,inner sep=0mm] {}; \&
            \node[rotate=0, align=center,inner sep=0mm] {\sf\scriptsize Input}; \&
            \node[rotate=0, align=center,inner sep=0mm] {\sf\scriptsize After Reflectance\\[-2mm]\sf\scriptsize and Illumination}; \&
            \node[rotate=0, align=center,inner sep=0mm] {\sf\scriptsize After Texture}; \&
            \node[rotate=0, align=center,inner sep=0mm] {\sf\scriptsize After Geometry}; \\
            \node[rotate=90, inner sep=0mm] {}; \&
            \node {\includegraphics[width=20mm]{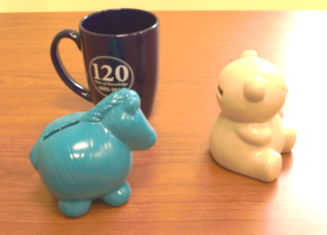}}; \&
            \node {\includegraphics[width=20mm]{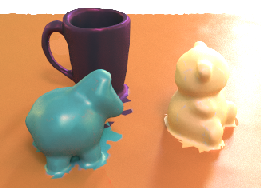}}; \&
            \node {\includegraphics[width=20mm]{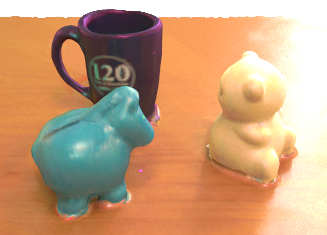}}; \&
            \node {\includegraphics[width=20mm]{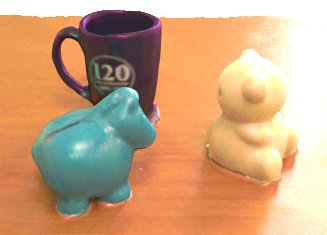}}; \\
        };
    \end{tikzpicture}
\caption{%
Stages of our method.
After reflectance, illumination, and segmentation converge (second column), the
method incorporates texture estimation (third column), and finally geometry
estimation (fourth column).
Each new step gives a progressively better result.
}
\vspace{-2mm}
\label{fig:results:stages}
\end{figure}

We have presented a method for decomposing a scene into its radiometric
elements (reflectance, illumination, and geometry) given images and rough
geometry of that scene from a small set of views.
Our method achieves reliable radiometric scene decomposition
by modeling and constraining complex reflectance and scene illumination,
and refining geometry while accounting for non-local light transport including
interreflection and shadowing.
The systematic evaluation on synthetic data and real-world scenes demonstrate
the effectiveness of our method.
The method enables the extraction of rich physically-based scene information
which may benefit many applications in computer vision and gives a new role to RGB-D sensors.

\section{Acknowledgements}
This work was supported by the Office of Naval Research grant N00014-16-1-2158 (N00014-14-1-0316) and the National Science Foundation awards IIS-1353235 and IIS-1421094.


\bibliographystyle{splncs}
\bibliography{mvrani_eccv16_paper}

\end{document}